\documentclass[conference]{IEEEtran}
\IEEEoverridecommandlockouts

\usepackage{cite}
\usepackage{amsmath,amssymb,amsfonts}
\usepackage{algorithmic}
\usepackage{graphicx}
\usepackage{textcomp}
\usepackage{xcolor}
\usepackage{orcidlink}
\usepackage{bm}
\usepackage{multirow}

\def\BibTeX{{\rm B\kern-.05em{\sc i\kern-.025em b}\kern-.08em
    T\kern-.1667em\lower.7ex\hbox{E}\kern-.125emX}}
\begin{document}

\title{High-Order Epistasis Detection Using Factorization Machine with Quadratic-Optimization Annealing and MDR-Based Evaluation
\thanks{This work was partially supported by the Japan Society for the Promotion of Science (JSPS) KAKENHI (Grant Number JP23H05447), the Council for Science, Technology, and Innovation (CSTI) through the Cross-ministerial Strategic Innovation Promotion Program (SIP), ``Promoting the application of advanced quantum technology platforms to social issues'' (Funding agency: QST), Japan Science and Technology Agency (JST) (Grant Number JPMJPF2221). }
}

\author{
    \IEEEauthorblockN{Shuta Kikuchi \orcidlink{0009-0002-1319-6834}}
    \IEEEauthorblockA{
        \textit{Graduate School of Science and Technology,} \\
        \textit{Keio University Sustainable Quantum} \\
        \textit{Artificial Intelligence Center (KSQAIC),} \\
        \textit{Keio University} \\
        Kanagawa, Japan \\
        kikuchi.shuta@keio.jp
    }
    \and
    \IEEEauthorblockN{Shu Tanaka \orcidlink{0000-0002-0871-3836}}
    \IEEEauthorblockA{
        \textit{Department of Applied Physics and Physico-Informatics,} \\
        \textit{Graduate School of Science and Technology,} \\
        \textit{Keio University Sustainable Quantum} \\
        \textit{Artificial Intelligence Center (KSQAIC),} \\
        \textit{Human Biology-Microbiome-Quantum} \\
        \textit{Research Center (WPI-Bio2Q),} \\
        \textit{Keio University} \\
        Kanagawa, Japan \\
        shu.tanaka@keio.jp
    }
}

\maketitle

\begin{abstract}
Detecting high-order epistasis is a fundamental challenge in genetic association studies due to the combinatorial explosion of candidate locus combinations. 
Although multifactor dimensionality reduction (MDR) is a widely used method for evaluating epistasis, exhaustive MDR-based searches become computationally infeasible as the number of loci or the interaction order increases. 
In this paper, we define the epistasis detection problem as a black-box optimization problem and solve it with a factorization machine with quadratic-optimization annealing (FMQA).
We propose an efficient epistasis detection method based on FMQA, in which the classification error rate (CER) computed by MDR is used as a black-box objective function. 
Experimental evaluations were conducted using simulated case--control datasets with predefined high-order epistasis. 
The results demonstrate that the proposed method successfully identified ground-truth epistasis across various interaction orders and the numbers of genetic loci within a limited number of iterations. 
These results indicate that the proposed method is effective and computationally efficient for high-order epistasis detection.
\end{abstract}


\section{Introduction}
Feature selection is important in many application areas involving high-dimensional data, since it simultaneously improves prediction accuracy, reduces computational and experimental costs, and facilitates a deeper understanding of the underlying processes that generate the observed data~\cite{guyon2003introduction}.
In biology and life science research, feature selection has been widely used as a key methodology for identifying disease-associated molecular features and candidate diagnostic biomarkers from high-dimensional biomedical data, such as omics, genomic, and clinical datasets~\cite{saeys2007review, torres2019research}.

Epistasis detection is one of the feature selection problems in biological and biomedical research.
Epistasis is defined as the interaction between multiple genetic loci, in which their combined effect influences a phenotype, such as disease status.
Detecting epistasis can provide deeper insights into the genetic architecture of complex traits and diseases. 
This is because it reveals non-additive effects that single-locus analyses do not capture, thereby contributing to explaining missing heritability.
Furthermore, identifying epistasis can enhance disease risk prediction and facilitate biological interpretation by highlighting groups of loci that act together within genetic networks or pathways.
Recent evidence indicates that high-order interactions involving more than two loci contribute to the genetic architecture of complex traits and diseases~\cite{taylor2014genetic, taylor2015higher}.
Therefore, effective methods to detect high-order epistasis among a large number of genetic loci are required.

Many methods have been proposed to detect epistasis~\cite{niel2015survey, balvert2024considerations}.
Multifactor dimensionality reduction (MDR) is one of the representative and widely used methods for epistasis detection, especially for higher-order interactions~\cite{ritchie2001multifactor}.
MDR entails reducing the high‐dimensional multi‐locus information into a one‐dimensional variable by partitioning multi‐locus genotypes into high‐risk and low‐risk groups.
However, standard MDR relies on exhaustive evaluation of all possible $d$-order combinations of genetic loci.
Therefore, the computational costs rapidly increase as the number of loci or the interaction order increases~\cite{shang2011performance, niel2015survey}.
Several studies have sought to preserve the strengths of MDR while mitigating its computational limitations.
For example, fast MDR (FMDR) adopts a greedy search strategy based on joint effects to reduce the number of evaluated combinations~\cite{yang2015efficiency}.
In addition, Crush-MDR integrates domain knowledge such as gene and pathway information to guide stochastic searches within the MDR~\cite{moore2017grid}.
Although such MDR-based acceleration methods substantially reduce the computational cost, they still face fundamental limitations.
Most existing approaches rely on heuristic search strategies, such as greedy or stochastic exploration, or require external domain knowledge to guide the search.
As a result, their performance may be sensitive to search biases, parameter settings, or the availability of prior knowledge, and the exploration of high-order epistatic interactions remains difficult.

In this paper, we propose a novel method in which a factorization machine with quadratic-optimization annealing (FMQA) is used to efficiently search for epistasis candidates, and MDR is employed to evaluate their classification performance.

FMQA is a black-box optimization method proposed by Kitai et al.~\cite{kitai2020designing}.
Originally, FMQA was referred to as factorization machine with quantum annealing.
The FMQA algorithm uses a factorization machine (FM)~\cite{rendle2010factorization} as a surrogate model, and optimizes the acquisition function using Ising machines.
Ising machines have attracted attention as fast and high-precision solvers for combinatorial optimization problems~\cite{mohseni2022ising}, operating with various internal algorithms~\cite{mohseni2022ising, kikuchi2025effectiveness} such as simulated annealing (SA)~\cite{kirkpatrick1983optimization} and quantum annealing (QA)~\cite{kadowaki1998quantum}.
FMQA has been applied across a wide range of domains~\cite{tamura2025black}, including material design~\cite{kitai2020designing, nawa2023quantum}, engineering design~\cite{inoue2022towards, matsumori2022application}, and peptide design~\cite{tucs2023quantum}.
It has been reported to identify combinations achieving better objective values with fewer function evaluations than Bayesian optimization and genetic algorithms~\cite{kitai2020designing, inoue2022towards, nawa2023quantum}.
Therefore, FMQA has the potential to substantially reduce the number of MDR evaluations required for epistasis detection.

The rest of this paper is organized as follows. 
Section~\ref{sec:method} reviews the preliminary methodologies, FMQA and MDR, and introduces the proposed method.
Section~\ref{sec:experimental_evaluation} describes the experimental setup and presents results and discussion.
Section~\ref{sec:conclusion} presents the conclusions and future work. 

\section{Method}
\label{sec:method}
In this section, we introduce FMQA and MDR.
After that, we describe the proposed method.

\subsection{FMQA}
\label{subsec:fmqa}
In this subsection, we introduce the flow of FMQA.
\begin{description}
   \item[Step 1] Initial training data are prepared, consisting of randomly generated solutions of binary variables $\bm{x} = (x_1, \dotsc , x_N) \in \{0, 1\}^N$ and the corresponding costs of the black-box (BB) function $y=f(\bm{x})$.
   Let $D$ denote the dataset constructed from these data.
   \item[Step 2] FM is trained with dataset $D$. The FM is defined by
    \begin{equation}
        E_{\rm{FM}}=\omega_{0}+\sum_{i=1}^{N}\omega_{i}x_{i}+\sum_{1\le i < j \le N}\langle \bm{v}_i, \bm{v}_j \rangle x_{i}x_{j},
        \label{eq:H_FM}
    \end{equation}
    where $\langle \bm{v}_i, \bm{v}_j \rangle$ is defined by
    \begin{equation}
        \langle \bm{v}_i, \bm{v}_j \rangle = \sum_{k=1}^{K} {v}_{i, k} {v}_{j, k}.
    \end{equation}
    Here, $\omega_0$, $\{\omega_{i}\}_{i=1,\cdots,N}$, and $\{v_{i,k}\}_{i=1,\cdots,N, k=1,\cdots,K}$ are the model parameters.
    The symbol $\langle \cdot, \cdot \rangle$ denotes the inner product.
    The parameter $K \in \mathbb{N}$ is the only hyperparameter of the FM.
   \item[Step 3] Using an Ising machine, new solutions are obtained by optimizing the acquisition function defined by the trained FM.
   We employ the predicted cost of the FM directly as the acquisition function.
   Ising machines search for better solutions to combinatorial optimization problems formulated by an Ising model or its equivalent model called a quadratic unconstrained binary optimization (QUBO).
   The Hamiltonian of the QUBO model is defined by:
    \begin{align}
      E_{\rm{QUBO}} = \sum_{1\leq i \leq j \leq N}Q_{i, j}x_{i}x_{j},
      \label{eq:H_QUBO}
    \end{align}
    where $x_{i} \in \{0, 1\}$ and $Q_{i, j}$ is the ($i$, $j$)-th element of the $N$-by-$N$ QUBO matrix $Q$.
    The FM model in~\eqref{eq:H_FM} can be converted into a QUBO formulation.
    \begin{equation}
        Q_{i,j} =
    \begin{cases}
        \omega_i, & (i = j), \\
        \langle \bm{v}_i, \bm{v}_j \rangle, & (i \neq j),
    \end{cases}
    \label{eq:fm_qubo}
    \end{equation}
    where $x_{i}^2 = x_{i}$ for any $i$.
    Therefore, the FM can be solved using Ising machines.
    The constant term $\omega_0$ is omitted in the QUBO formulation, as it does not affect the optimization result.
    \item[Step 4] A BB function is called to evaluate the actual costs for the solutions obtained in Step $3$. 
    Then, pairs of solutions and their corresponding costs are appended to the dataset $D$.
    \item[Step 5] Steps $2$--$4$ are repeated for a predefined number of iterations.
    After completion, the solution with the lowest cost in the dataset $D$ is selected as the best solution obtained by FMQA.
\end{description}

\subsection{MDR}
\label{subsec:mdr}
\begin{figure}[t]
    \centering
    \includegraphics[clip,width=0.68\linewidth]{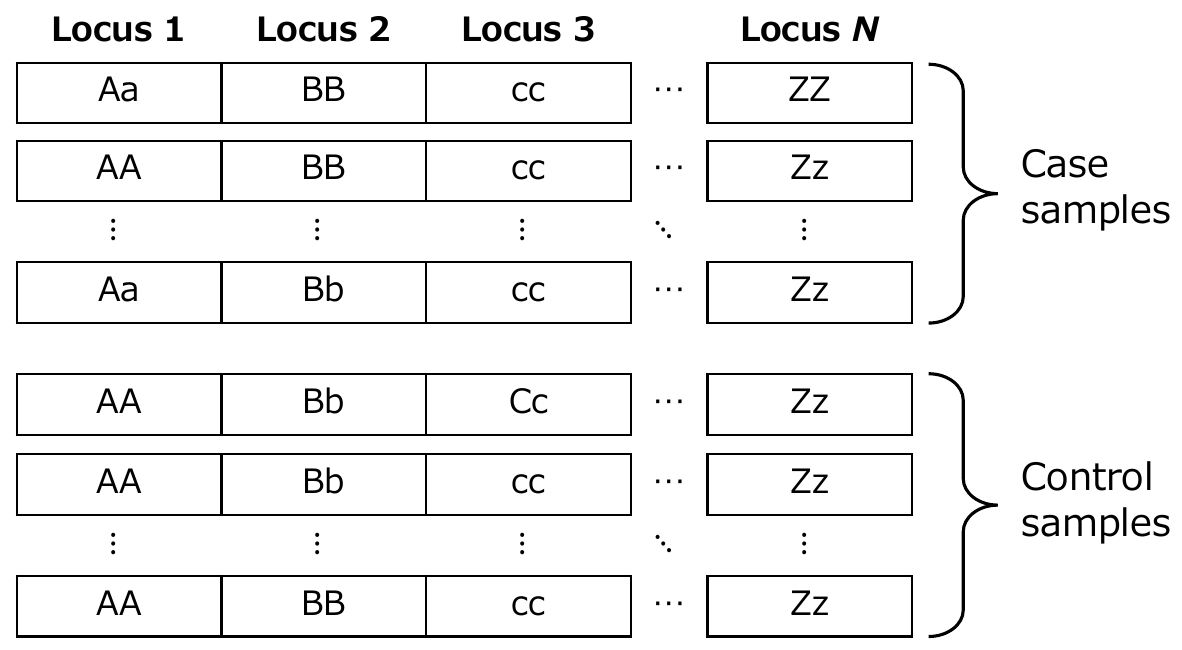}
    \caption{An example of the case--control dataset.
    Each column corresponds to a genetic locus, and each entry represents the genotype of an individual at that locus.
    Genotypes are denoted using standard diploid notation, each locus has three possible genotypes, where uppercase and lowercase letters indicate different alleles (e.g., AA and aa denote homozygous genotypes, and Aa denotes a heterozygous genotype with one copy of each allele).
    Each row corresponds to one individual sample, which is labeled as either a case or a control according to the phenotype.
    }
    \label{fig:dataset}
\end{figure}

In this subsection, we describe the flow of MDR.
MDR was originally proposed by Ritchie et al.~\cite{ritchie2001multifactor} as a non-parametric and model-free method for epistasis detection.
In this study, we adopt an MDR formulation applicable to case--control datasets with imbalanced case--control ratios~\cite{velez2007balanced, yang2013mdr}.
Here, \textit{cases} refer to samples exhibiting the target phenotype
(e.g., disease) and \textit{controls} refer to samples without the target phenotype.
A case--control dataset contains genotype information at each locus for samples classified as either cases or controls.
Fig.~\ref{fig:dataset} illustrates an example of the case--control dataset used in the evaluation of MDR.
\begin{description}
    \item[Step 1] The case--control dataset is divided into $z$ subsets for $z$-fold cross-validation (typically $z=10$).
    For each fold, one subset is used as an independent test set, while the remaining $z-1$ subsets are combined to form the training set.
    \item[Step 2] For a fixed order $d$, all possible combinations of $d$ loci are exhaustively enumerated from the $N$ available loci in the case--control dataset.
    \item[Step 3] For each $d$-locus combination, the training data are mapped into a contingency table, where each cell corresponds to a specific multi-locus genotype.
    The numbers of cases and controls assigned to each cell are counted.
    For the $i$-th multi-locus genotype cell, the normalized ratio of cases to controls $\theta_i$ is defined as
    \begin{equation}
        \theta_i = \dfrac{P_i / N_i}{P^* / N^*},
    \label{eq:f_L}
    \end{equation}
    where $P^*$ and $N^*$ denote the numbers of cases and controls in the training set, respectively.
    Here, $P_i$ and $N_i$ represent the numbers of cases and controls, respectively, that exhibit the $i$-th multi-locus genotype.
    Each cell is labeled as \textit{high-risk} (H) if $\theta_i$ exceeds a predefined threshold $T (=1)$, and as \textit{low-risk} (L) otherwise.
    Note that the denominator in Eq.~\eqref{eq:f_L} can be zero when no control samples fall into a genotype cell.
    If a cell contains at least one case but no controls, $\theta_i$ is treated as infinity and the cell is labeled as high-risk.
    If neither cases nor controls fall into the cell, we set $\theta_i = 0$ and label it as low-risk.
    \item[Step 4] By pooling all high-risk genotype cells into one group and all low-risk cells into another, the original $d$-dimensional genotype information is reduced to a single binary attribute.
    Using this binary attribute, a $2 \times 2$ contingency table (TP, FP, TN, FN) is constructed for the training set.
    Here, true positives (TP) and false negatives (FN) correspond to case samples classified as high-risk and low-risk, respectively, whereas false positives (FP) and true negatives (TN) correspond to control samples classified as high-risk and low-risk.
    \item[Step 5] The classification performance of each $d$-locus model is evaluated using the classification error rate (CER) defined as
    \begin{equation}
        \mathrm{CER} = \frac{1}{2}
        \left(
        \frac{\mathrm{FN}}{\mathrm{TP}+\mathrm{FN}} +
        \frac{\mathrm{FP}}{\mathrm{FP}+\mathrm{TN}}
        \right).
    \end{equation}
    \item[Step 6] For each cross-validation fold, the $d$-locus model with the minimum CER in the training set is recorded to compute the cross-validation consistency (CVC), and its prediction error is evaluated on the test set. 
    The final epistatic model is selected as the $d$-locus combination with the highest CVC and the lowest average prediction error.
\end{description}

\subsection{Proposed method}
\label{subsec:proposed_method}
\begin{figure*}[t]
    \centering
    \includegraphics[clip,width=1.0\linewidth]{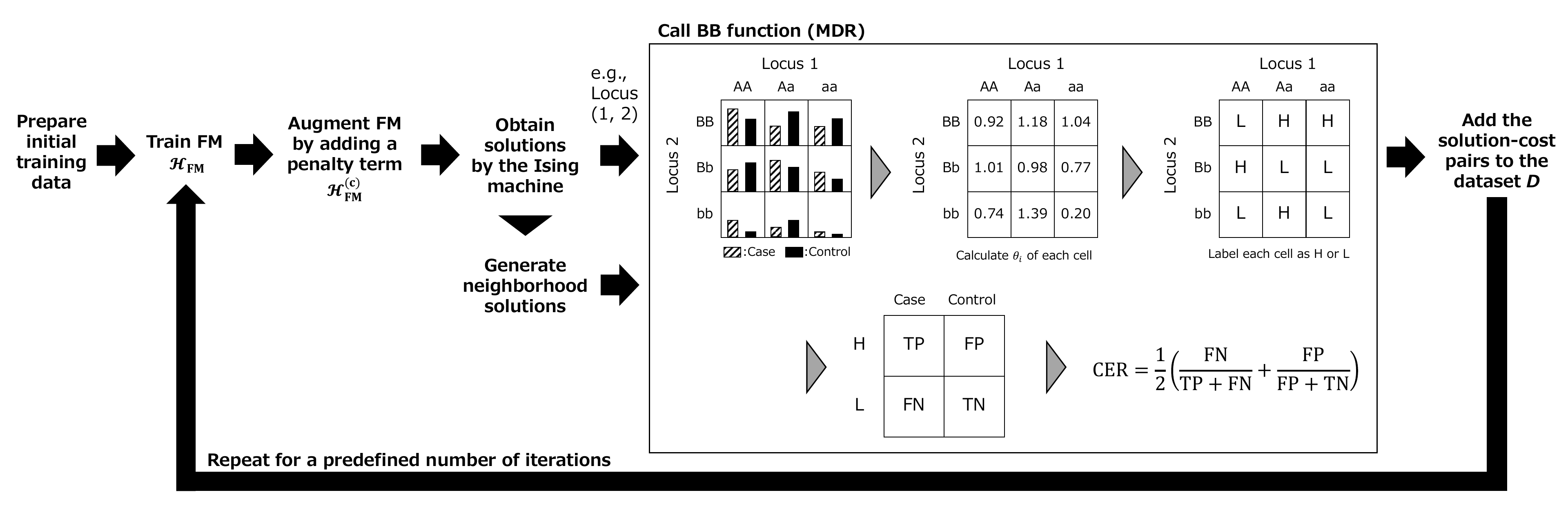}
    \caption{Flow of the proposed method. The black arrows and gray triangles denote the flow for the proposed method and for CER computation using MDR, respectively. This figure illustrates an example with $d=2$, where the solution obtained by the Ising machine proposes an epistasis candidate between locus~$1$ and locus~$2$ that affects the case status.
    }
    \label{fig:FMQA-MDR_procedure}
\end{figure*}

In this subsection, we present a novel framework for epistasis detection that integrates FMQA and MDR.
The idea of the proposed method is to use MDR as the BB function in FMQA, thereby enabling efficient identification of informative $d$-locus epistasis without exhaustive search.
Fig.~\ref{fig:FMQA-MDR_procedure} shows the workflow of the proposed method.

Let $N$ denote the total number of genetic loci. 
We introduce a binary vector $\bm{x} = (x_1, x_2, \ldots, x_N) \in \{0,1\}^N$, where $x_i = 1$ indicates that the $i$-th locus is selected as an epistasis candidate and $x_i = 0$ otherwise. 

The objective of the proposed method is to identify the $d$-locus epistasis combination that minimizes the CER.
To focus on $d$-locus epistasis, a constraint is added through a penalty term in the FM Hamiltonian $E_{\mathrm{FM}}$ introduced in Section~\ref{subsec:fmqa}.
This results in the following constrained formulation:
\begin{equation}
    E_{\mathrm{FM}}^{\mathrm{(c)}}(\bm{x})
    =
    E_{\mathrm{FM}}(\bm{x})
    +
    \lambda
    \left(
    \sum_{i=1}^{N} x_i - d
    \right)^2 ,
\end{equation}
where $\lambda > 0$ is a penalty coefficient that controls the strength of the constraint.

An Ising machine is used to obtain solutions by minimizing $E_{\mathrm{FM}}^{\mathrm{(c)}}$. 
Based on each solution obtained from the Ising machine, additional neighborhood solutions are generated by perturbing the original solutions.
Then, the costs (CER) corresponding to these solutions are evaluated, and the solution--cost pairs are appended to the dataset $D$.
This approach is motivated by a previous study~\cite{matsumori2022application}, which demonstrated that evaluating neighborhood solutions enables FMQA to obtain high-accuracy solutions with a limited number of iterations.
When the solution obtained by the Ising machine satisfies the constraint $\sum_i x_i = d$, neighborhood solutions are generated using a swap-based operation, in which one variable with $x_i = 1$ is flipped to $0$ and another variable with $x_j = 0$ is flipped to $1$.
This operation preserves the constraint while enabling exploration of nearby combinations of genetic loci.
To avoid generating infeasible solutions, a sufficiently large penalty coefficient $\lambda$ was used, and no infeasible solutions were observed in our experiments.

Unlike the standard MDR procedure described in Section~\ref{subsec:mdr}, cross-validation is not employed in this evaluation. 
We note that computing CER on all samples may increase the risk of overfitting and false-positive epistasis.
However, the purpose of this study is to evaluate the efficiency of the FMQA-based search under the MDR-defined objective, rather than to assess the statistical significance or generalization performance of the detected models.
Therefore, we compute CER using all available samples in the case--control dataset.
For each $d$-locus combination represented by a solution obtained from the Ising machine or its neighborhood solutions, Steps $3$--$5$ described in Section~\ref{subsec:mdr} are applied, and the corresponding CER is used as the cost for each solution.

\section{Experimental evaluation}
\label{sec:experimental_evaluation}
In this section, we describe the experimental setup and present results and discussion.

\begin{table*}[t]
    \caption{Performance of the proposed method for each instance.
    The success rate denotes the ratio of runs in which the ground-truth epistasis was identified. 
    Avg. iter. represents the average number of FMQA iterations required to reach a successful solution, computed only over successful runs.
    Max. iter. denotes the maximum number of iterations observed among successful runs.
    }
    \label{tab:results}
    \centering
    \begin{tabular}{cc|ccc|ccc|ccc}
    \hline
    \multirow{2}{*}{Order $d$} & \multirow{2}{*}{Model} 
    & \multicolumn{9}{c}{Number of attributes $N$} \\
    \cline{3-11}
    & 
    & \multicolumn{3}{c|}{100}
    & \multicolumn{3}{c|}{500}
    & \multicolumn{3}{c}{1000} \\
    \cline{3-11}
    & 
    & Success rate & Avg. iter. & Max. iter.
    & Success rate & Avg. iter. & Max. iter.
    & Success rate & Avg. iter. & Max. iter. \\
    \hline
    \multirow{2}{*}{3}
    & Additive  & 1.0 & 34.4 & 40  & 1.0 & 311.5 & 350 & 1.0 & 587.6 & 799 \\
    & Threshold & 1.0 & 36.2 & 42  & 1.0 & 281.1 & 303 & 1.0 & 578.8 & 600 \\
    \hline
    \multirow{2}{*}{4}
    & Additive  & 1.0 & 28.2 & 41  & 1.0 & 293.1 & 369 & 1.0 & 553.4 & 770 \\
    & Threshold & 1.0 & 52.4 & 95  & 1.0 & 293.3 & 441 & 1.0 & 559.5 & 744 \\
    \hline
    \multirow{2}{*}{5}
    & Additive  & 1.0 & 41.6 & 145 & 1.0 & 243.6 & 315 & 1.0 & 657.0 & 803 \\
    & Threshold & 1.0 & 90.0 & 141 & 0.7 & 315.1 & 468 & 0.9 & 633.8 & 757 \\
    \hline
    \end{tabular}
\end{table*}

\subsection{Setup}
In this subsection, we describe the procedure for generating the simulated case--control datasets used for evaluation and the parameter settings of the proposed method.

The experimental evaluation was conducted using simulated case--control datasets with predefined epistasis in order to quantitatively assess the performance of the proposed method.

To generate the datasets, first, penetrance tables representing high-order epistasis were constructed using Toxo~\cite{ponte2020toxo}.
Toxo is a penetrance-table generator that calculates epistasis models of arbitrary interaction order by fixing either the heritability or the prevalence and maximizing the other under the constraints imposed by a specified epistatic model.
In this study, we considered two representative epistasis models provided by Toxo: the \emph{additive} model and the \emph{threshold} model.
The additive model represents epistatic interactions in which the phenotypic effect increases cumulatively with the number of loci contributing to the phenotype, resulting in both marginal effects and interaction effects.
In contrast, the threshold model assigns an elevated probability of the target phenotype only when specific combinations of genetic loci exceed a predefined threshold.
This emphasizes interaction effects over the marginal contributions of individual loci.
These two models represent qualitatively different genetic architectures.
The difficulty of the generated epistasis models can be characterized in terms of marginal effects.
The additive model corresponds to epistasis displaying marginal effects (eME), whereas the threshold model can be regarded as representing epistasis displaying no marginal effects (eNME).
These two categories are widely used to distinguish between detectable and challenging epistatic scenarios.
eNME is considered more difficult to detect than eME.
This is because each locus involved in the ground-truth epistasis does not exhibit a strong marginal signal, and even $d$-locus combinations that include a subset of the true loci are unlikely to produce a strong signal.

After generating penetrance tables, case--control datasets were generated using GAMETES~\cite{urbanowicz2012gametes}.
GAMETES uses the supplied penetrance tables to simulate case--control datasets while preserving the specified minor allele frequencies (MAFs) and heritability values. 

In the experiments, we considered high-order epistasis of order $d = 3, 4,$ and $5$. 
The MAF for all causal loci was set to $0.4$, and the heritability was fixed at $0.2$. 
The total number of genetic loci (the number of attributes) was set to $N = 100, 500,$ and $1000$ in order to evaluate scalability with respect to problem dimensionality.
For each dataset, the number of samples was fixed at $1000$ cases and $1000$ controls.

Next, we describe the parameter settings used for the proposed method.
For the FM, the hyperparameter $K$ was set to $10$, following previous studies~\cite{inoue2022towards, nakano2026swift}.
The number of initial training data points was fixed to $10$.
Each initial binary vector $\bm{x}$ was constructed so that exactly $d$ elements were set to $1$, while the remaining elements were set to $0$.
The indices of the active variables were selected uniformly at random.
We employed the Fixstars Amplify Annealing Engine~\cite{FixAE}, version 0.9.0, as the SA-based Ising machine, which runs on a GPU.
We set the time limit to 1 second per Ising solve, and all other solver hyperparameters (e.g., temperature schedule and number of sweeps) were kept at the defaults.
The Step-3 backend in FMQA can be replaced by alternative optimizers, including QA hardware or QAOA-based solvers. 
Evaluating such replacements is left for future work.
The coefficients of the FM Hamiltonian $E_{\rm FM}$ were first normalized by the maximum absolute value of the coefficients.
Then, a penalty term with a coefficient $\lambda = 2$ was added to enforce the constraint throughout the experiments.
With this setting, all solutions obtained by the Ising machine satisfied the constraint $\sum_{i=1}^N x_i = d$ in our experiments.
At each FMQA iteration, a single solution was obtained by the Ising machine, and one neighborhood solution was generated.
When generating the neighborhood solution, the variables to be swapped were randomly selected.
The total number of FMQA iterations was set to $1000$.
For each problem instance, the proposed method was performed $10$ times using different randomly generated initial training data.
A trial was considered successful if the ground-truth epistasis appeared at least once among the solutions evaluated during the FMQA iterations within the maximum number of iterations.
Here, the success rate reflects the robustness of the FMQA-based search with respect to different random initializations.

\subsection{Results and discussion}
In this subsection, we present the results of the experimental evaluation of the proposed method.
Table~\ref{tab:results} summarizes the performance of the proposed method for each instance.
Table~\ref{tab:results} indicates the success rate, defined as the ratio of runs in which the ground-truth epistasis was successfully detected.
A solution that correctly identifies the ground-truth epistasis is referred to as a successful solution.
In addition, the table shows the average number of iterations required to obtain a successful solution, as well as the maximum number of iterations observed among the successful solutions.

Table~\ref{tab:results} demonstrates that the proposed method is capable of identifying the ground-truth epistasis across all instances considered in this evaluation.
When the number of attributes was $100$, successful solutions were detected with fewer than $100$ iterations on average for all considered interaction orders and models.
However, when the interaction order increased to $d=5$ under the threshold model, the average number of iterations increased.
For the cases with $500$ attributes, successful solutions were detected within approximately $300$ iterations on average for most instances.
In contrast, when the order was five and the threshold model was used, the proposed method failed to obtain successful solutions within $1000$ iterations in three runs.
When the number of attributes was increased to $1000$, successful solutions were detected within approximately $600$ iterations on average.
For $d = 5$, the required number of iterations was larger compared to lower-order interactions.
In addition, for the threshold model with $d = 5$, the proposed method failed to obtain a successful solution in one run.
We conducted uniform random search for all instances using the same total number of evaluations (2000 evaluations).
Under this setting, random search failed to identify the ground-truth epistasis in any instance.

These results indicate that the number of iterations required to obtain a successful solution increases as the number of attributes grows.
This trend can be attributed to the expansion of the candidate solution space, which increases the difficulty of efficiently identifying the true epistasis.
However, Table~\ref{tab:combination_order}, which shows the number of possible combinations when selecting $d$ loci from $N$ $(_N C_d)$, indicates that the number of iterations required for the proposed method to obtain a successful solution is not directly determined by the order of the solution space.
Interestingly, the number of iterations does not scale directly with the solution space size shown in Table~\ref{tab:combination_order}, and instead increases more moderately as $N$ grows.

In addition, for higher interaction orders, the threshold model tended to require a larger number of iterations than the additive model and in some runs failed to yield successful solutions.
Moreover, in several runs, the ground-truth epistasis was not obtained within the predefined maximum number of iterations.
This can be attributed to the fact that the threshold model corresponds to eNME, in which marginal effects are absent or very weak. 
Under the eNME setting, combinations that include only a subset of the true epistatic loci rarely achieve low CER values, because marginal and lower-order effects are absent.
As a result, informative intermediate solutions that could guide the FM toward the true epistasis are scarce, making it difficult for the FM to learn a surrogate that assigns low costs to the correct high-order combination.

\begin{table}[t]
    \caption{Number of combinations required for exhaustive search for different
    numbers of attributes $N$ and orders $d$.}
    \label{tab:combination_order}
    \centering
    \begin{tabular}{c|c|c|c}
    \hline
    Order $d$ & \multicolumn{3}{c}{Number of attributes $N$} \\
    \cline{2-4}
     & 100 & 500 & 1000 \\
    \hline
    3 & $1.6 \times 10^{5}$ & $2.1 \times 10^{7}$ & $1.7 \times 10^{8}$ \\
    4 & $3.9 \times 10^{6}$ & $2.6 \times 10^{9}$ & $4.1 \times 10^{10}$ \\
    5 & $7.5 \times 10^{7}$ & $2.6 \times 10^{11}$ & $8.3 \times 10^{12}$ \\
    \hline
    \end{tabular}
\end{table}

\section{Conclusion and future work}
\label{sec:conclusion}
In this paper, we proposed an FMQA-based framework for efficiently identifying candidate high-order epistasis under the MDR criterion, in which the CER computed by MDR is used as a BB function.
It should be noted that the objective of this study is to evaluate the ability of the proposed FMQA-based framework to efficiently explore $d$-locus epistasis candidates by minimizing the MDR-based classification error rate computed on the full dataset.
Accordingly, the present evaluation focuses on search efficiency and does not address the generalization performance or statistical reproducibility of the detected epistatic models.
By defining the epistasis detection problem as a black-box optimization problem and solving it with FMQA, the proposed method avoids exhaustive enumeration of all $d$-locus combinations while maintaining high detection performance.
Experimental evaluations using simulated case--control datasets demonstrated that the proposed method successfully identified ground-truth epistasis across a wide range of problem settings, including high-order interactions and large numbers of attributes.
These results indicate that the proposed method is effective and computationally efficient for high-order epistasis detection.
Moreover, the proposed method has the potential to be extended to other biomedical feature selection problems, such as biomarker discovery.

Several directions remain for future investigation.
First, the robustness of the proposed method should be further evaluated using more difficult case--control datasets by varying parameters such as the MAF and heritability.
Second, future studies should examine statistical variability by conducting experiments on a larger number of independent dataset instances.
Finally, applying the proposed method to real-world genetic datasets will be essential to assess its practical applicability.

\section*{Acknowledgment}
S. Tanaka wishes to express his gratitude to the World Premier International Research Center Initiative (WPI), MEXT, Japan, for their support of the Human Biology-Microbiome-Quantum Research Center (Bio2Q).
The computations in this work were partially performed using the facilities of the Supercomputer Center, the Institute for Solid State Physics, The University of Tokyo.

\bibliographystyle{IEEEtran}
\bibliography{ref}

\end{document}